# A New Color Feature Extraction Method Based on Dynamic Color Distribution Entropy of Neighborhoods

**Fatemeh Alamdar[1] and MohammadReza Keyvanpour[2]**

**[1] Department of Computer Engineering, Alzahra University
Tehran, Iran**

**[2] Department of Computer Engineering, Alzahra University
Tehran, Iran**

## Abstract

One of the important requirements in image retrieval, indexing, classification, clustering and etc. is extracting efficient features from images. The color feature is one of the most widely used visual features. Use of color histogram is the most common way for representing color feature. One of disadvantage of the color histogram is that it does not take the color spatial distribution into consideration. In this paper dynamic color distribution entropy of neighborhoods method based on color distribution entropy is presented, which effectively describes the spatial information of colors. The image retrieval results in compare to improved color distribution entropy show the acceptable efficiency of this approach.

## 1. Introduction

In recent decades, digital technology progress results in unprecedented growth in production of digital images. Therefore development of effective automatic techniques for image sets organization and management is required so that one can search, retrieval and categorize the images more convenient. Feature extraction is the basis of these automatic techniques. Color, texture and shape are the most common visually features[1]. These features are independent of specific domain and can used in general systems of retrieval images[2]. The color feature is the first and one of the most widely used visual features in image retrieval and indexing[3]. The most important advantages of color feature are power of representing visual content of images, simple extracting color information of images and high efficiency, relatively power in separating images from each other[4], relatively robust to background complication and independent of image size and orientation[5,6,2].

The color histogram method introduced in [7] has shown to be very effective and simple to implement. Use of color histogram is the most common way for representing color feature[8]. Despite of some drawbacks, color histogram had been used in many researches and great efforts were done for overcoming its weakness [9-14]. One of disadvantage of the color histogram method is that it is not robust to significant appearance changes because it does not include any spatial information[10]. Several schemes including spatial information have been proposed. Pass et al. [15] suggested classifying each pixel as coherent or no coherent based on whether the pixel and its neighbors have similar colors. Then, a split histogram called color coherence vector (CCV) is used to represent this classification for each color in an image. Huang[11] proposed a color correlograms method, which collects statistics of the co-occurrence of two colors some distances apart. A simplification of this feature is the autocorrelogram, which only captures the spatial correlation between identical colors. In [11-14] respectively introduced annular color histogram, spatial-chromatic histogram (SCH) and geostat to describe how pixels of identical color are distributed in the image. Sun et al. [10] propose a color distribution entropy (CDE) method, which takes account of the correlation of the color spatial distribution in an image. This feature is based on annular color histogram that draws some concentric circles from images and then the annular color histogram is calculated by counting the pixels of every color bin inside every circle. The number of circles is a predefined constant and for every image is the same regardless of its content.

In this paper we introduce a dynamic color distribution entropy of neighborhoods (D-CDEN) method, which similar to CDE describes the spatial information of an image. Instead of drawing concentric circles, D-CDEN takes account of images' content by attending to neighborhoods of pixels for every color bin. The number of extracted neighborhoods is different for every color bins. In addition, predefining this number does not require in this approach, also the color indexing results in image clustering and retrieval is much better. The results are demonstrated by image retrieval and D-CDEN in compare to I-CDE show the efficiency of this approach.

The rest of the paper is organized as follows. A briefly review on CDE and I-CDE is presented in Section 2. Section 3 describes the proposed feature extraction base on D-CDEN. Section 4 details the similarity measurement. Experimental results are demonstrated in Section 5. Finally, a conclusion is given in Section 6.

## 2. Color Distribution Entropy

CDE descriptor was proposed in [10]. This descriptor expresses the color spatial information of an image. This descriptor based on the NSDH (Normalized Spatial Distribution Histogram) and information entropy was defined. NSDH was derived from Annular Color Histogram[12]. In Annular Color Histogram which introduced by Rao et al., suppose $A_i$ be the set of pixels with color bin $i$ of an image and $|A_i|$ be the number of elements in $A_i$. Let $C_i$ be the centroid and $r_i$ be the radius of color bin $i$ which are defined in [12]. With $C_i$ as the center and with $jr_i / N$ as the radius for each $1 \le j \le N$, $N$ concentric circles can be drawn. Let $|A_{ij}|$ be the count of the pixels of color bin $i$ inside circle $j$. Then the annular color histogram can be written as $(|A_{i1}|, |A_{i2}|, ...., |A_{iN}|)$. This is illustrated in Fig. 1. Based on the Annular Color Histogram, the NSDH is given in Eq. (1).

$$P_i = (P_{i1}, P_{i2}, ..., P_{iN})$$
$$P_{ij} = |A_{ij}| / |A_i|$$
(1)

The $E_i$ defined as CDE of color bin $i$, was defined as

$$E_i(P_i) = \sum_{j=1}^{N} P_{ij} \log_2(P_{ij})$$
(2)

where $P_i$ is the Normalized Spatial Distribution Histogram.

This equation shows the dispersive degree of the pixel patches of a color bin in an image. Large $E_i$ means the distribution of the pixels is dispersed, otherwise the distribution is compact. Then the CDE index for an image can be written as $(h_1, E_1 ... h_i, E_i, ..., h_n, E_n)$, where $h_i$ is the histogram of color bin $i$, $E_i$ is the CDE of color bin $i$ and $n$ is the number of bins.

The improved CDE (I-CDE) was defined as

$$E_i(P_i) = -g(P_i) \sum_{j=1}^{N} f(j) P_{ij} \log_2(P_{ij})$$
(3)

$$f(j) = 1 + \frac{j}{N}$$
(4)

$$g(P_i) = 1 + \frac{A(P_i)}{N}$$
(5)

$$A(P_i) = \sum_{j=1}^{N} (P_{ij} \times j)$$
(6)

$f(i)$ is the weight function which denotes the different contribution of each annular circle to the CDE. $g(P_i)$ is the weight function using Histogram Area( $A(P_i)$ ) defined as Eq. (6). $g(P_i)$ effectively removes the influence of symmetrical property of entropy. More details could be found in [10].

The CDE and I-CDE similarity measurement of image $I_q$ and $I_t$ was defined as[10]

$$d(I_q, I_t) = \sum_{i=1}^{n} \min(h_i^{I_q}, h_i^{I_t}) \times \frac{\min(E_i^{I_q}, E_i^{I_t})}{\max(E_i^{I_q}, E_i^{I_t})}$$
(7)

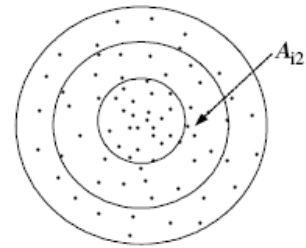

Fig 1: Annular Color Histogram[10]

### 2.1 Footnotes

Footnotes should be typed in singled-line spacing at the bottom of the page and column where it is cited. Footnotes should be rare.

## 3. Feature extraction based on Dynamic Color Distribution Entropy of Neighborhoods

D-CDEN method is based on CDE and effectively describes the spatial information of colors. In CDE, two images are considered similar when distributions of the pixels of color bins are the same but layout and neighborhoods of color pixels can be not the same, so this distribution may be similar in different images. D-CDEN method takes account of images' content and instead of drawing $N$ concentric circles in CDE attends to neighborhoods of pixels for every color bin of image color histogram.

## 3.1 Neighborhoods extraction

For extracting neighborhoods for every color bin $i$, an image matrix is scanned rows by rows from left to right, up to down. Because of this kind of scanning, only neighborhoods of right and up adjacent pixels of current pixel had been identified and regarded. If none of them is in the same color bin, this pixel is in the new neighborhood; but in the other cases, if the current pixel is in the same color bin of the right, $135°$ diagonal or up adjacent pixels, it is assigned to its neighborhood. It is illustrated in Fig. 2. If the middle pixel is the current pixel, the 1-8 pixels are its neighbors (Fig. 2(a)). Because this pixel and pixel 2 are in the same color bin, it is in the neighborhoods of pixel 2. In Fig. 2(b) the neighborhoods which were detected up to current pixels are determined by different numbers.

A problem may be appear in the neighborhoods' specifying when the current pixel is in the same color bin of both right and up adjacent pixels but their neighborhoods are different. In this case these two neighborhoods are merged. Fig. 3 shows this problem. Pixel 0 is the current pixel and is not in the same color bin of 1,2,3 pixels (Fig. 3(a)) and Fig. 3(b) shows detected neighborhoods up to now, so a new neighborhood is defined for current pixel. In continues of scan, when the last pixel is the current pixel, this problem occur (Fig. 3(c)). Fig. 3(d) shows the final detected neighborhoods when neighborhood 1, 7 had been merged.

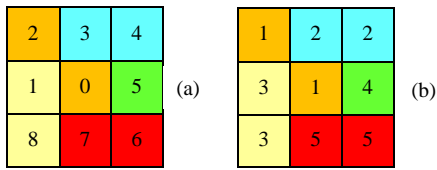

Fig. 2: (a) neighbor pixels of pixel 0. (b) extracted neighborhoods for image (a)

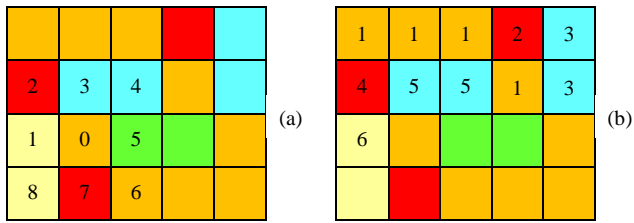

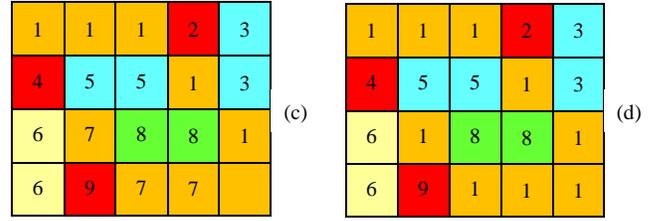

Fig. 3: (a) neighbor pixels of pixel 0. (b,c,d) extracted neighborhoods for image (a)

Extracted neighborhoods for a real image are shown in Fig. 4(b). In this figure, neighborhoods of a color bin have the same color.

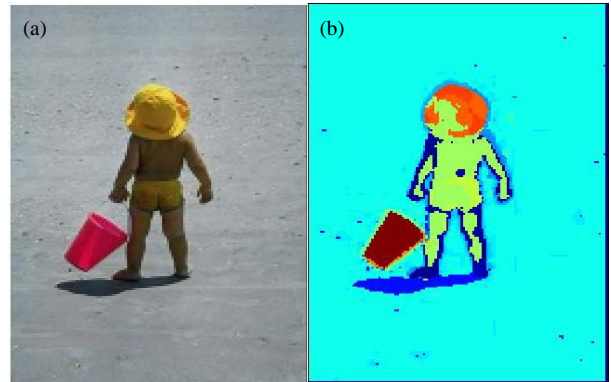

Fig. 4: (a) original image. (b) neighborhoods of image (a).

## 3.2 Dynamic Color Distribution Entropy of Neighborhoods

Like CDE, Normalized Spatial Distribution Histogram is defined as:

$$\boldsymbol{P}_i' = (P_{i1}', P_{i2}', \ldots, P_{inb_i}')$$
$$\text{and } P_{ij}' = |\boldsymbol{A}_{ij}'| / |\boldsymbol{A}_i'|$$
$$\text{for } 1 \le j \le nb_i \qquad (8)$$

$nb_i$ is the number of extracted neighborhoods for color bin $i$ and is different for every color bins and $\boldsymbol{A}_i'$ is the set of pixels with color bin $i$ of an image and $|\boldsymbol{A}_{ij}'|$ is the count of the pixels of neighborhood $j$ for color bin $i$.

The D-CDEN is defined as Eq. (2) by replacing $N$ with $nb_i$ and $\boldsymbol{P}_i$ with $\boldsymbol{P}_i'$, so the D-CDEN index for an image is written as $(h_1, E_1', \ldots h_i, E_i', \ldots, h_n, E_n')$, where $E_i'$ is the D-CDEN of color bin $i$.

## 3.3 Feature extraction

Firstly, before extraction of neighborhoods, the images are resized into 128x128 pixels, because it makes noises removed and small neighborhoods reduced especially in cluttered scene images. Then for every image, neighborhoods are extracted and $nb_i$ is calculated according the previous sub-sections. Afterwards the images are indexed by D-CDEN descriptor and in HSV color space. The color space is uniformly quantized into 8 levels of hue, 2 levels of saturation and value giving a total of 32 bins.

## 4. Similarity measurement

In this case, we introduce a dissimilarity measurement based on vector space retrieval model (VSM), which was used in [16].

In [16], two problems of CDE similarity measurement have been mentioned. First problem is that two color bins are similar or two color bins are not similar but just have the same histogram, so different images with the same histogram are considered similar. Another problem is that the same $d(I_q, I_t)$ can be produced by a number of different sets. In order to overcome these problems, the similarity measurement was done by using vector space retrieval model (VSM). VSM measure is as follows[17]:

$$\cos\theta_H(I_q, I_t) = \frac{\sum_{i=1}^{n} h_i^{t_q} \times h_i^{t_t}}{\sqrt{\sum_{i=1}^{n} h_i^{t_q \, 2}} \times \sqrt{\sum_{i=1}^{n} h_i^{t_t \, 2}}} \qquad (9)$$

$$\cos\theta_E(I_q, I_t) = \frac{\sum_{i=1}^{n} E_i^{t_q} \times E_i^{t_t}}{\sqrt{\sum_{i=1}^{n} E_i^{t_q \, 2}} \times \sqrt{\sum_{i=1}^{n} E_i^{t_t \, 2}}} \qquad (10)$$

$$d'(I_q, I_t) = 2 - (\cos\theta_H(I_q, I_t) + \cos\theta_E(I_q, I_t)) \qquad (11)$$

where $\cos\theta_H(I_q, I_t)$ and $\cos\theta_E(I_q, I_t)$ are the color histogram similarity and color distribution entropy similarity between two images $I_q$ and $I_t$ in vector space retrieval model, and $d'(I_q, I_t)$ is the dissimilarity of two images.

For measuring the image dissimilarity, we use the following distance:

$$d''(I_q, I_t) = 3 - (\cos\theta_H(I_q, I_t) + \cos\theta_E(I_q, I_t) + \cos\theta_N(I_q, I_t)) \qquad (12)$$

where $\cos\theta_H(I_q, I_t)$ and $\cos\theta_E(I_q, I_t)$ are calculated by Eq. (9) and (10) by replacing $E$ with $E'$ and $\cos\theta_N(I_q, I_t)$ is computed as:

$$\cos\theta_N(I_q, I_t) = \frac{\sum_{i=1}^{n} Nb_i^{t_q} \times Nb_i^{t_t}}{\sqrt{\sum_{i=1}^{n} Nb_i^{t_q \, 2}} \times \sqrt{\sum_{i=1}^{n} Nb_i^{t_t \, 2}}} \qquad (13)$$

where $Nb_i$ is the normalized number of neighborhoods for color bin $i$ for an image, which is defined as:

$$Nb_i = \frac{nb_i}{\sum_{j=1}^{n} nb_j} \qquad (14)$$

## 5. Experimental Results

In this section, the results of D-CDEN are demonstrated by image retrieval and these results are compared with I-CDE. Experiments were carried out by using two databases. The first is SIMPLIcity[1][18] database of 1000 images which included 10 categories of Africa people, Beach, Buildings, Buses, Dinosaurs, Elephants, Flowers, Horses, Mountains and Food. Every category contains 100 images. Fig. 5 shows the different types of images in this experimental database. Each of the images is with the size of $256 \times 384$ or $384 \times 256$ pixels.

Secondly we used 70 categories of Caltech101 [2] [19] database. Categories contain of 6384 images of different objects. There are about 40 to 800 images in each object category. Size of each image is roughly $200 \times 300$ or $300 \times 200$ pixels. Some of the different types of images in this database are shown in Fig. 6.

For comparing, an image query was chosen and the retrieval results of using on D-CDEN, I-CDE has been calculated and for specifying the number of circles in I-CDE, we averaged over $nb_i$s mean $\left(\frac{1}{n}\sum_{i=1}^{n} nb_i\right)$ for all of images. The used image dissimilarity measurements for D-CDEN and I-CDE are $d''$ and $d'$ respectively. The retrieval accuracy was measured in terms of the Recall, Precision. The Precision rate and Recall rate are defined as follows:

---

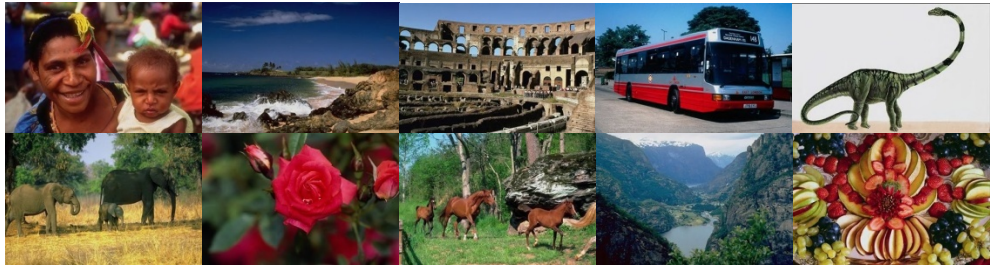

Fig. 5: Different types of image in SIMPLIcity database

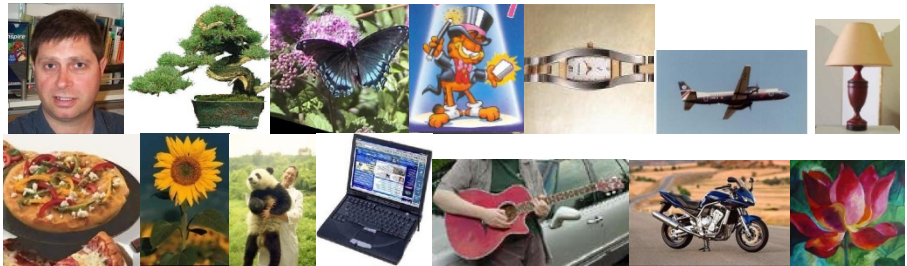

Fig. 6: Different types of image in Caltech101 database

$$Precision = \frac{r}{Nr}$$

$$Recall = \frac{r}{Ni}$$

(15)

where $r$ is the number of relevant images selected, $Nr$ is the total number of retrieved images and $Ni$ is the total number of similar images in the database. Results show the superiority of D-CDEN.

For comparing in SIMPLIcity database, every 100 of the categories was chosen as a query image. Fig. 7 is the Recall and Precision graph of results averaged over 100 images in Buildings(7(a)), Buses(7(b)), Flowers(7(c)).

In Caltech101 database, 50 images were selected randomly as query images. Fig. 8 is the Recall and Precision graph of results averaged over 50 random selected queries in 3 times.

Because of attending to neighborhoods of color pixels and the number of them, D-CDEN has better results in both databases.

In Fig. 9, the top-left one (9(a)) is the query image from SIMPLIcity database and the top ten retrieval results of query are sorted by similarity from left-to-right and top-to-down sequence by using of D-CDEN method(9(b)) and I-CDE method(9(b)). The same results for Caltech101 database are shown in Fig. 10.

## 6. Conclusions

In this paper, a color features extraction method based on dynamic color distribution entropy of neighborhoods was expressed. D-CDEN method measures the spatial relation of colors in an image and takes account of images' content by neighborhoods extraction of pixels for every color bin of image color histogram.

In this work we introduce a new dissimilarity measuring to demonstrating results by image retrieval and these results are compared with I-CDE. Experiments were carried out by using two databases of 1000 and 6384 images. These experiments show the acceptable efficiency of this approach.

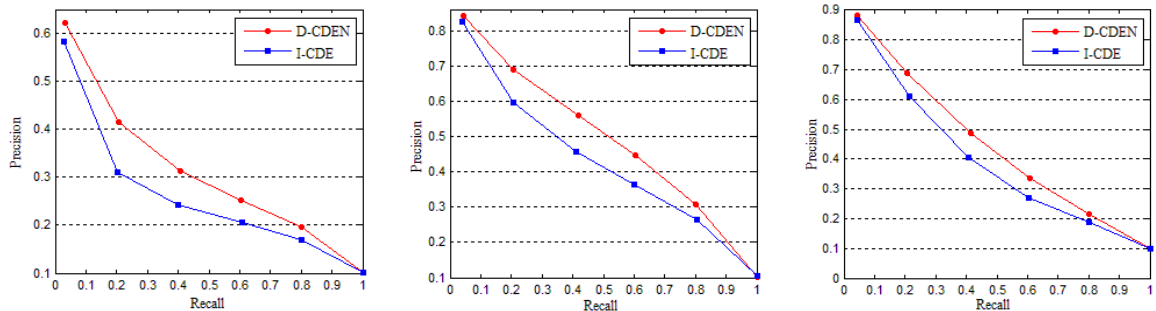

Fig. 7: Precision/Recall graph

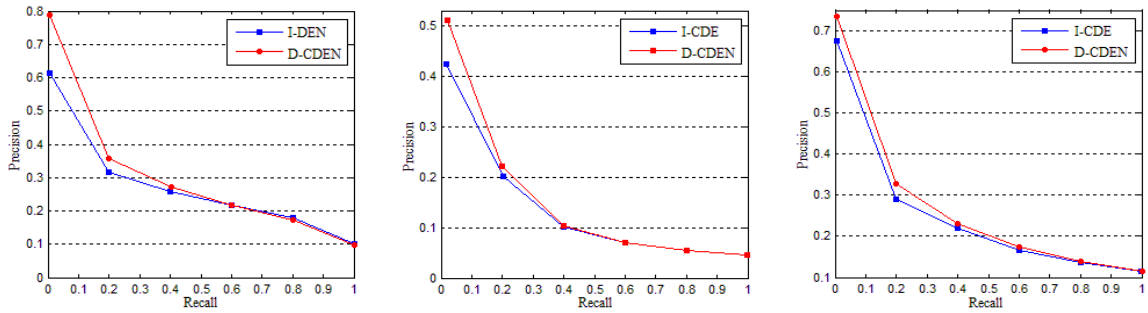

Fig. 8: Precision/Recall graph

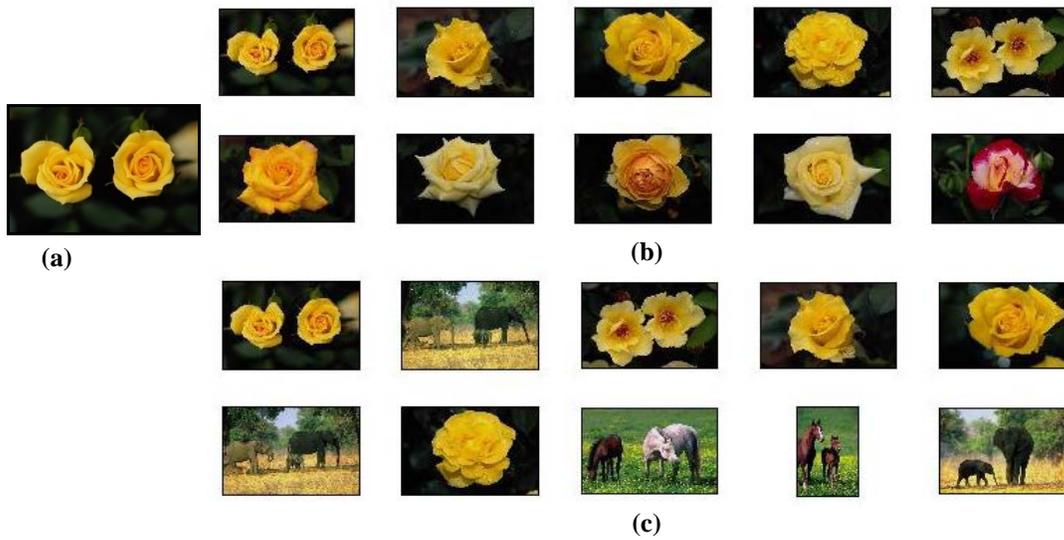

Fig. 9: (a) Query image from SIMPLIcity database. (b) Query results of D-CDEN method. (c) Query results of I-CDE method

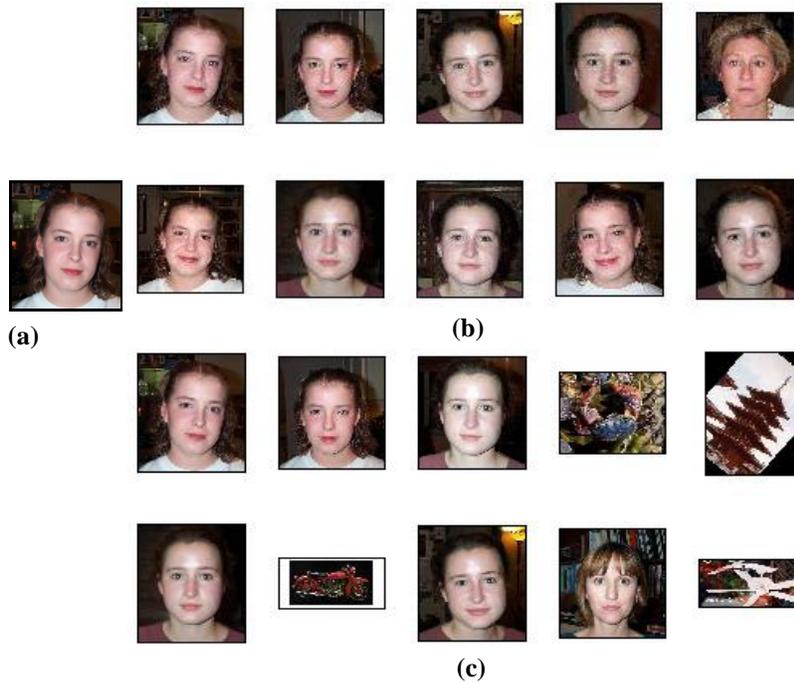

Fig. 10: (a) Query image from Caltech101 database. (b) Query results of D-CDEN method. (c) Query results of I-CDE method

## Acknowledgments

This work was supported in part by the Iran Telecommunication Research Center, ITRC.

## References

[1] L. Hove, "Improving Content Based Image Retrieval Systems with a Thesaurus for Shapes", M.S. Thesis, Institute for Information and Media Sciences, University of Bergen, Bergen, Norway, 2004.

[2] M. Keyvanpour, N. Moghadam Charkari, "Image Retrieval Using Hybrid Visual Features", in ICEE, Tehran, Iran, 2008, pp. 62-67.

[3] R. S. Torres, A. X. Falcao, "Content-Based Image Retrieval: Theory and Applications", Revista de Informática Teórica e Aplicada, Vol. 13, No.2, 2006, pp.161-185.

[4] S. Panchanathan, Y. Park, et al., "The Role of Color in Content-Based Image Retrieval", in ICIP'00, Canada, 2000, Vol. 1, pp. 517-520.

[5] Y. Rui and T.S. Huang, "Image retrieval: Current techniques promising directions and open issues", Journal of Visual Communication and Image Representation, Vol.10, 1999, pp. 39-62.

[6] J. Zhang, W. Hsu, M. Lee, "An Information driven Framework for Image Mining", in Proc of 12th International Conference on Database and Expert Systems Applications, Munich, Germany, 2001, pp. 232-242.

[7] M. J. Swain, D. H. Ballard, "Color indexing, International Journal of Computer Vision", Vol. 7, No. 1, 1991, pp. 11-32.

[8] L. Tran, "Efficient Image Retrieval with Statistical Color Descriptors", Ph.D. Thesis, Department of Science and Technology, Linkoping University, Linkoping, Sweden, 2003.

[9] R. Datta, D. Joshi, J. Li, and J.Z. Wang, "Image Retrieval: Ideas, Influences, and Trends of the New Age", ACM Computing Surveys, Vol. 40, No. 2, 2008, pp. 1-60.

[10] J. Sun, X. Zhang, J. Cui and L. Zhou, "Image retrieval based on color distribution entropy", Pattern Recognition Letters, Vol. 27, No. 10, 2006, pp. 1122-1126.

[11] J. Huang, "Image indexing using color correlograms", in IEEE Computer Society Conference on Computer Vision and Pattern Recognition. San Juan., 1997, pp. 762–768.

[12] A. B. Rao, R. K. Srihari, and Z. F. Zhang, "Spatial color histogram for content-based retrieval, Tools with artificial intelligence", in Proceedings of 11th IEEE International Conference, 1999, pp. 183-186.

[13] L. Cinque, S. Levialdi, K. Olsen, et al., "Color-based image retrieval using spatial-chromatic histogram", in IEEE International Conference on Multimedia Computing and Systems, 1999, Vol. 2, pp. 969-973.

[14] S. Lim, G.J. Lu, "Spatial statistics for content based image retrieval", in International Conference on Information Technology: Computers and Communications, Las Vegas, Nevada, 2003, pp. 28-30.

[15] G. Pass, R. Zabih, J. Miller, "Comparing images using color coherence vectors", in ACM 4th International Conference on Multimedia, Boston, Massachusetts, United States, 1996, pp. 65-73.

[16] G. Lui, B. Lee, "A Color-based Clustering Approach for Web Image Search Results", in International Conference on Convergence and Hybrid Information Technology, Korea, 2009, pp. 481-484.

[17] G. Lu, multimedia database management systems, USA: Artech house publisher, 1999, pp. 81-82.

[18] J.Z. Wang, J. Li, and G. Wiederhold, "SIMPLIcity: semantic sensitive integrated matching for picture libraries", IEEE Trans. Pattern Anal. Machine Intell., Vol. 23, No. 9, 2001, pp. 947-963.

[19] L. Fei-Fei, R. Fergus, and P. Perona, "Learning generative visual models from few training examples: an incremental Bayesian approach tested on 101 object categories", CVPR 2004, Workshop on Generative-Model Based Vision, 2004.